\documentclass[graybox]{svmult}

\usepackage{mathptmx}		
\usepackage{helvet}			
\usepackage{courier}			
\usepackage{type1cm}		
\usepackage{makeidx}		
\usepackage{graphicx}		
\usepackage{multicol}        	
\usepackage[bottom]{footmisc}	
\usepackage{color}
\usepackage{microtype}
\usepackage{caption}
\usepackage{subcaption}
\usepackage{url}

\usepackage[breaklinks]{hyperref}
\usepackage{breakurl}

\makeindex             

\begin{document}

\title*{Communicating Robot Arm Motion Intent Through Mixed Reality Head-mounted Displays}
\titlerunning{Robot Arm Motion Intent in Mixed Reality}

\author{Eric Rosen\thanks{First two authors contributed equally}, David Whitney$^{*}$, Elizabeth Phillips, Gary Chien,\\James Tompkin, George Konidaris, Stefanie Tellex}

\authorrunning{Rosen et al.}

\institute{Eric Rosen, David Whitney, and Stefanie Tellex \at Humans To Robots Lab, Brown University. \{eric\_rosen,david\_whitney,stefanie\_tellex\}@brown.edu
\and Elizabeth Phillips \at Humanity Centered Robotics Initiative, Brown University, elizabeth\_phillips1@brown.edu
\and Gary Chien and James Tompkin \at Brown University, \{gary\_chien,james\_tompkin\}@brown.edu
\and George Konidaris \at Intelligent Robot Lab, Brown University, george\_konidaris@brown.edu
}

%
%
\maketitle

\abstract{
Efficient motion intent communication is necessary for safe and collaborative work environments with collocated humans and robots. Humans efficiently communicate their motion intent to other humans through gestures, gaze, and social cues. However, robots often have difficulty efficiently communicating their motion intent to humans via these methods. Many existing methods for robot motion intent communication rely on 2D displays, which require the human to continually pause their work and check a visualization. We propose a mixed reality head-mounted display visualization of the proposed robot motion over the wearer's real-world view of the robot and its environment. To evaluate the effectiveness of this system against a 2D display visualization and against no visualization, we asked 32 participants to labeled different robot arm motions as either colliding or non-colliding with blocks on a table. We found a 16\% increase in accuracy with a 62\% decrease in the time it took to complete the task compared to the next best system. This demonstrates that a mixed-reality HMD allows a human to more quickly and accurately tell where the robot is going to move than the compared baselines.
}

\section{Introduction}
\label{sec:introduction}
Industrial robots excel at performing precise, accurate, fast, and repetitive tasks. This makes them ideal for activities like car assembly. One major drawback of these robots is that humans are unable to easily predict their motions, which forces most industrial robots to be isolated from human workers and restricts human-robot collaboration. This is especially true in a fluid working environment without rigidly-defined tasks, or where robots have autonomy. Although the intended robot motion is defined ahead of time through motion planning, efficiently conveying the intended motion to a human is difficult. Human-robot collaboration requires robots to communicate to humans in ways that are intuitive and efficient \cite{fong2003survey}; yet, the motion intention inference problem leads to many safety and efficiency issues for humans working around robots \cite{confused}. 

This problem has inspired research into how robots might effectively communicate intent to humans. Current interfaces for communicating robot intent have limitations in expressing motion plans within a shared workspace. Humanoid robots can try to mimic the gestures and social cues that humans use with each other, but many robots are not and cannot be humanoid by design. The motion robots intend to make can also be visualized on a 2D display near the robot. This requires the human to take their attention away from the robot's physical space to observe the display, which could be dangerous. Additionally, a 2D projection of a 3D motion plan can take time for a human to understand, requiring interaction to inspect different points of view.

It has been speculated that natural communication might be achieved when humans can see a robot's future motion in the real world from their own point of view, via a head-mounted display \cite{Ruffaldi2016, scassellati2014human}. This would potentially increase safety and efficiency as the human no longer needs to divert their attention. Further, as the 3D motion plan would be overlaid in 3D space, it would also eliminate the need for human participants to make sense of 2D projections of 3D objects.

We experimentally test this idea with a system to enable humans to view a robot's intended motion via 3D graphics on a mixed reality (MR) head-mounted display (HMD)---the Microsoft HoloLens. This allows a participant to visualize the motion of the robot's arm in the real workspace before it moves, potentially preventing collisions with the human or with objects. As there is no existing open source HoloLens ROS integration within the robotics community, we will release the \emph{ROS Reality} package. This integrates HoloLens with the widely-used Unity game engine, and provides a URDF parser to quickly import robots into Unity. 

\begin{figure}[p]
    HoloLens visualization \hspace{0.26\columnwidth} View captured directly from HoloLens \\
    \includegraphics[width=\columnwidth]{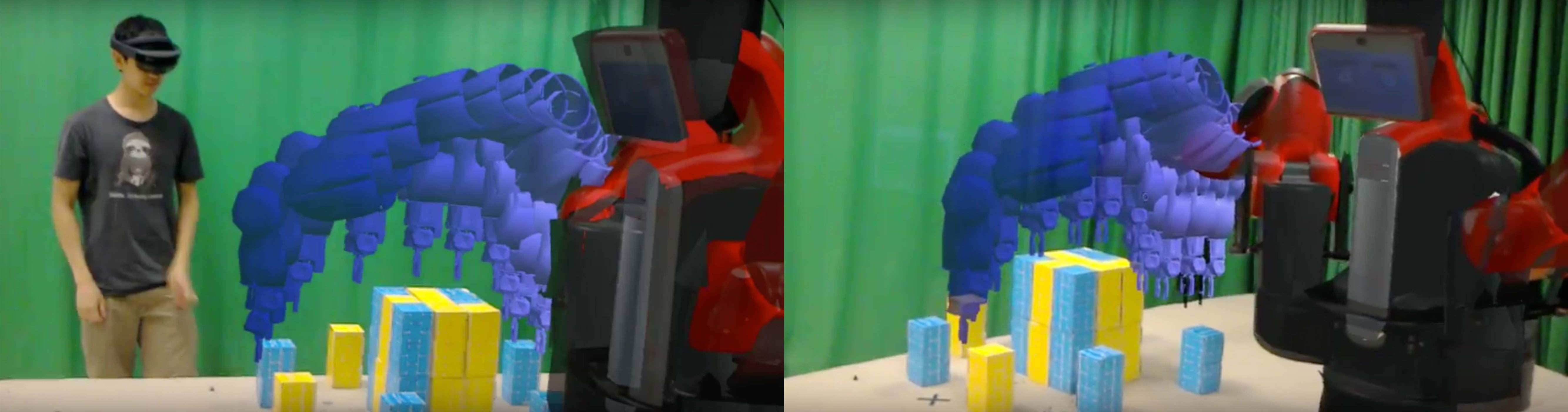} \\ \qquad \\
    2D display visualization \hspace{0.25\columnwidth} RViz-like interactive 3D scene \\
    \includegraphics[width=\columnwidth]{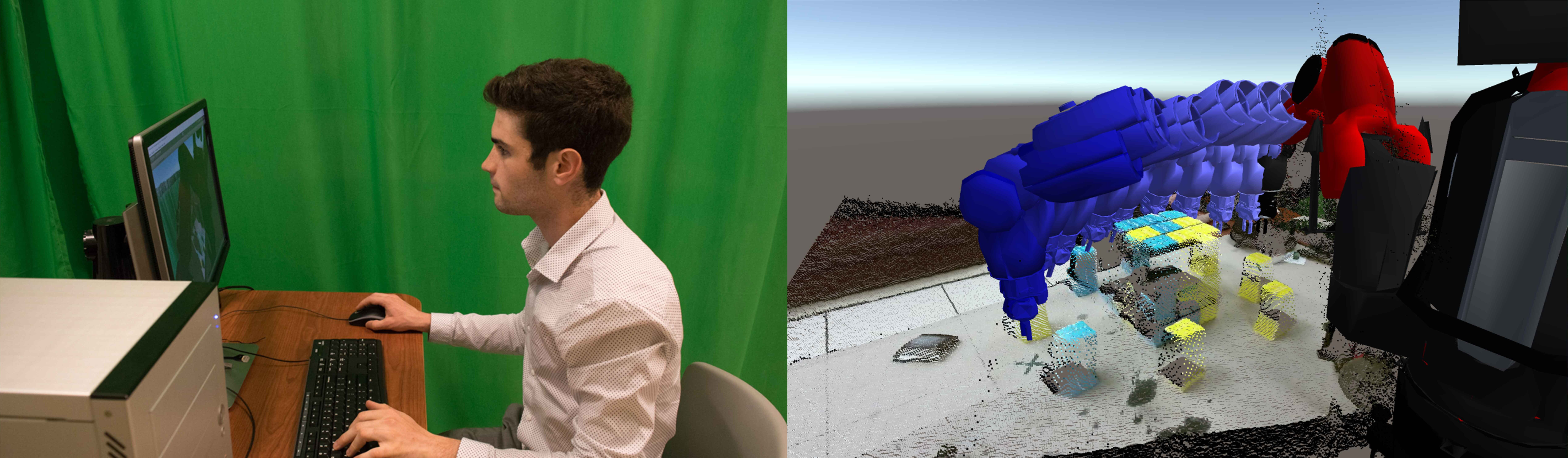} \\ \qquad \\
    No visualization \hspace{0.32\columnwidth} Stroboscopic photo of robot arm motion \\
    \includegraphics[width=.495\columnwidth]{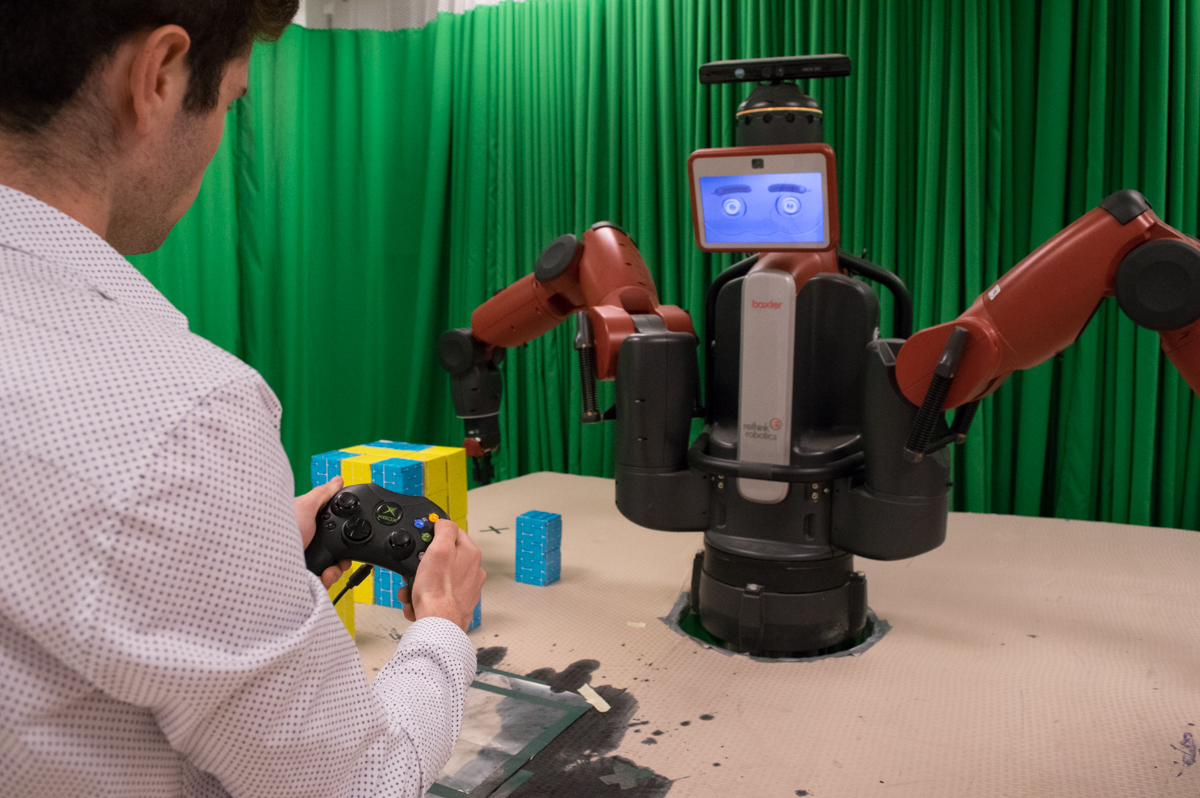}
    \includegraphics[width=.495\columnwidth]{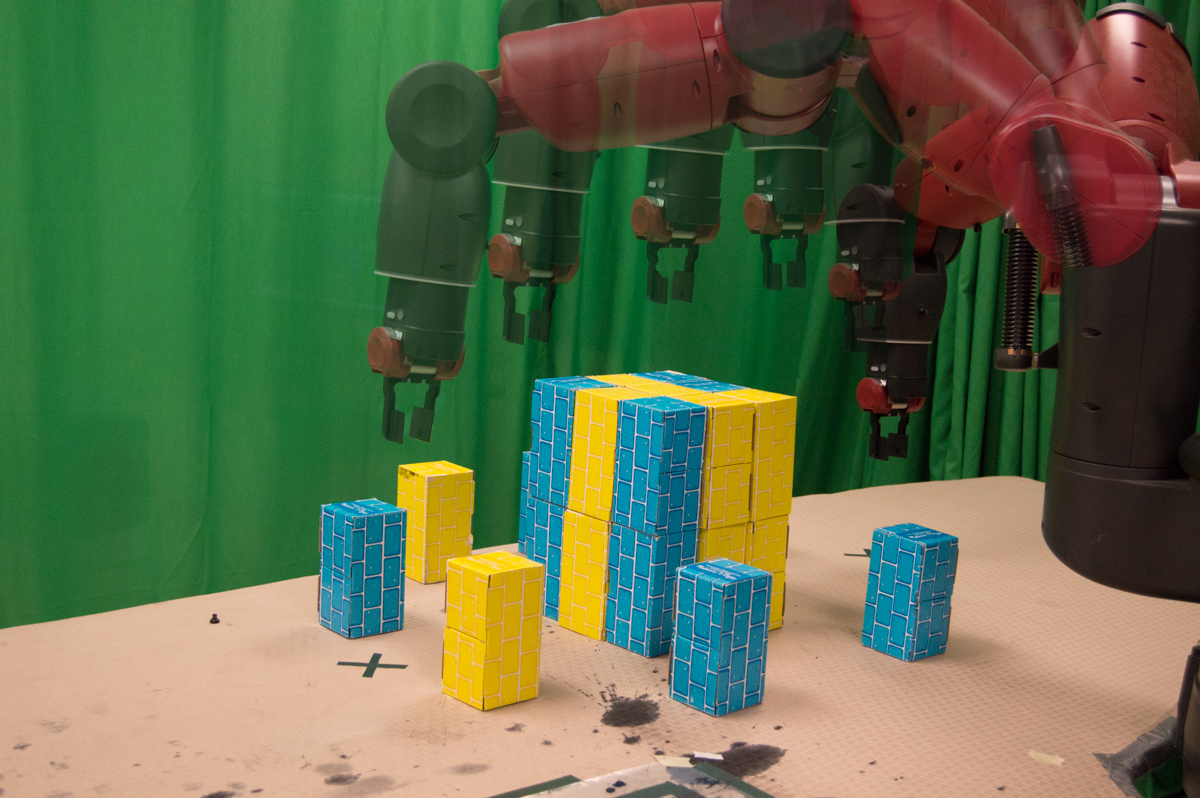}
    \caption{Participants must decide whether a robot arm motion plan either collides or not with the light yellow and blue blocks on the table, across 14 trials and three interfaces. \emph{Top to bottom:} Our three interfaces: a HoloLens visualization, a 2D display/mouse with an RVis-like visualization, and no visualization at all. In each case, the left shows the experimental setup, and the right shows the participant view. \emph{Top:} The HoloLens visualizes the robot arm motion plan as a sequence of blue virtual arm graphics overlaid onto the real world. \emph{Middle:} The 2D display uses the same visualization, but the participant must use the system at a desk. \emph{Bottom:} In the no visualization condition, the participant directly observes the robot arm move and pushes a `stop' button on an Xbox controller if they think collision will occur.}
	\label{fig:overview}
\end{figure}

We evaluated our MR system by comparing it to both a 2D display interface and a control condition with no visualization (Fig.~\ref{fig:overview}). In a within-subjects-design study, 32 participants used all three system variants to classify arm motion plans of a Rethink Robotics Baxter as either colliding or not colliding with blocks on a table. Our MR system reduced task completion time by 7.4 seconds on average (a reduction of 62\%), increased precision by 11\% percent on average, and increased accuracy by 16\% percent on average, compared to the next best system (2D display). Additionally, we improved subjective assessments of system usability (System Usability Scale) and mental workload (NASA Task Load Index). This experiment shows the promise of mixed-reality HMDs to further human-robot collaboration.

\section{Related Work}
\label{sec:relatedwork}
Humans use many non-verbal cues to communicate motion intent. There have been some successes at approximating these cues in humanoid robots, such as with gestures \cite{nakata1998expression} and gaze \cite{mutlu2009nonverbal}, including via robot anthropomorphism \cite{may2015show}. However, often robots lack the faculty or subtlety to physically reproduce human non-verbal cues---especially robots that are not of human form. One alternative is to use animation and animated storytelling techniques, such as forming suggestive poses or generating initial movements \cite{takayama2011expressing}. This increases legibility: the ability to infer the robot's goal through its directed motion \cite{dragan2013legibility}. However, these methods still lack the ability to transparently communicate complex paths and motions. Further, tasks involving close proximity teamwork may require more detailed knowledge of how the robot will act both before and during the motion, such as in collaborative furniture assembly \cite{scassellati2014human} and co-located teleoperation \cite{szafir2014communication}.

Other related works have used turn and display indicators on the robot to communicate navigational intent \cite{szafir2015communicating, chadalavada2016empirical, schaefer2017communicating}. These techniques were found to improve human trust and confidence in robot actions; however, they did not show a significant improvement in communicating high-fidelity navigational intent due to an inability to express high detail in the motion plan. \cite{shrestha2016exploring,shrestha2016intent}

We can also use 2D displays to visualize the robot's future motions within its environment through systems like RViz \cite{Kam2015}. However, these requires the human operator to switch focus from the real world environment to the visualization display \cite{milgram1993applications}. This may lead the operator to spend more time understanding the robot state and environment rather than collaborating with the robot \cite{burke2004moonlight,burke2004situation}. Often, the participant observes the environment through a potentially-noisy RGB video or point cloud feed, which adds uncertainty. 

\subsection{Augmented and Mixed Reality for Human-robot Collaboration}
\label{sec:relatedwork_ar}
We can adapt the real-world environment around the human-robot collaboration to help indicate robot intent. One way is to combine light projectors with object tracking software and virtual graphics to build a general-purpose augmented environment. This has been used to convey a shared work space, robot navigational intention, and safety information \cite{chadalavada2015s,ahn2016remote,andersen2016projecting}. However, building special purpose environments is time consuming and expensive, with limitations due to the occlusion of light from objects in the environment, on the number of people able to see perspective-correct graphics, and with a requirement for controlled lighting conditions.

Hand-held tablet technology can allow participants to view a mixed reality of 3D graphics overlaid onto a camera feed of the real world \cite{rekimoto1996transvision}. These types of approaches mediate the issue of diverted attention which 2D displays suffer. However, they limit the ability of the operator to use their hands while working, and there is a mismatch in perspective between the eyes of the human and the camera in the tablet.

Optical head-mounted displays can overlay 3D graphics on top of the real world from the point of view of the human. This has been speculated to be a natural and transparent means of robot intent communication, for instance, with the overlay of future robot poses \cite{Ruffaldi2016, scassellati2014human}. We imply this to mean that such a system would reduce human-robot collaborative task time and produce fewer errors. The recent introduction of the Microsoft HoloLens has made off-the-shelf implementations of such a visualization possible. Previously, the HoloLens and other similar MR interfaces have been used in human-human collaboration, such as communicating with remote companions and playing adversarial games \cite{kato1999marker, ohshima1998ar2, chen20153d}. However, mixed reality as a tool to communicate robot motion intent for human-robot collaboration is nascent. This inspired us to test the hypothesis that an MR HMD which allows participants to see visual overlays on top of real-world environment in human-robot collaborative tasks is more performant than existing approaches.

\subsection{3D Spatial Reasoning in Virtual Reality Displays}

As HoloLens and its contemporaries are relatively new as pieces of integrated technology, there is little direct evidence to support the speculated that optical HMDs will provide natural robot intent communication. However, hypotheses may be informed from literature in the parallel technology of virtual reality (VR) which, in a similar way to mixed reality (MR), provides head tracked stereo display of 3D graphics to create immersion. In VR, 3D spatial reasoning gains have been tested \cite{Slater2016}. Pausch et al.~found that head-tracked displays outperform stationary displays for a visual search task \cite{Pausch1993}. Ware and Franck find a head-tracked stereo display 3$\times$ less erroneous than a 2D display for visually assessing graph connectivity \cite{Ware1994}. Slater et al.~measured performance gains in Tri-D chess for first-person perspective VR HMDs over third-person perspective 2D displays (like RViz) \cite{slater1996immersion}. Ruddle et al.~found navigation through a 3D virtual building was faster using HMDs over 2D displays, though with no accuracy increase \cite{Ruddle1999}. 

Not all experiments in this area favor large-format VR. Many prior works compare immersive head-tracked CAVE displays against desktop and `fishtank VR' displays, and often smaller higher-resolution displays induce greater performance thanks to faster visual scanning \cite{Kasik2002,demiralp2006cave}. Sousa Santos et al.~reviewed all HMD to 2D display comparisons in the literature until 2009, and found their results broadly conflicting. Then, they conducted their own comparison for 3D navigation: on average, the desktop setup was better than the VR HMDs \cite{SousaSantos2009}.

In general, the relationship between VR display and task performance is one with many confounding factors. The benefits over traditional 2D desktop displays are task dependent, and no clear prescriptive guidelines exist for which techniques to employ to gain what benefit. As such, while we may assume that a mixed reality interface for viewing 3D would be better, the evidence from the VR literature tells us that the issue may be more complex.

\section{Experiment}
\label{sec:experiment}

To test whether mixed reality HMDs can aid robot motion intent communication, we asked novice participants to decide whether or not a robot arm motion plan would collide with blocks on a table using three interfaces: no visualization, an RViz-like 2D display visualization, and our HoloLens visualization. Our evaluation used 32 participants (15 male, 17 female) with ages ranging from 20 to 55 $\mbox(M = 26, SD = 6.8)$. We measure task completion time and true/false positive/negative rates as objective metrics, as well as the subjective assessments of system usability, likability, and workload via the SUS and NASA TLX questionnaires.

\subsection{Task}
In each interface, we presented each participant with the same set of 14 robot arm motions in a random order. These motions each moved from a fixed start point to a fixed end point over a table covered in blocks. Each participant was tasked with labeling the motions as either colliding or non-colliding with the blocks as quickly and accurately as possible. The blocks were assembled such that it would be difficult to obtain a complete view of all blocks from just one perspective due to occlusion from other blocks. Once a participant had decided how to classify a particular motion, they pressed a button on an Xbox controller to indicate their label. Participants could freely walk around to view the environment from different perspectives.

\subsection{Interfaces}
We compared three interfaces (Fig.~\ref{fig:overview}):

\begin{itemize}
\item \textbf{No visualization:}
This condition simulated a participant supervising a robot with an emergency stop button. Participants watched the arm move, and pressed an Xbox controller button to stop the arm if they thought it would collide.

\item \textbf{Monitor:}
Participants viewed and interacted with a 2D monitor on a desk. The visualization consisted of: 1) a 3D model of the robot and its future arm poses, obtained by importing a description of the robot in URDF format with its preplanned path joint poses, and 2) a 3D point-cloud of the environment, captured by a Kinect v2 sensor mounted near the robot. The robot's future arm poses changed color from a light to dark blue to represent what point in time the robot would be in that arm pose, allowing participants to see the entire planned path. In this interface, the robot arm did not move. Participants could move the virtual camera in the visualization to gain different perspectives using a keyboard-and-mouse-based control scheme (as in RViz \cite{Kam2015}). For consistency, participants recorded their assessment using an Xbox controller. 

\item \textbf{Mixed Reality (MR):}
Through HoloLens, participants viewed a similar 3D visualization of the robot with the motion overlaid on top of the real world. In this case, there is no need to visualize the environment via a point cloud because the participant can see it directly. Based on the visualization and their physical motion in the real world, participants decided upon whether the motion collided or not and recorded their prediction using an Xbox controller. Likewise, in this interface, the robot arm did not move.

\end{itemize}

\subsection{Experimental Procedure}
We began by reading a consent document to the participant. After consenting to participate, participants were asked to complete our navigational intent task using all three interfaces. The no visualization condition was always completed before the other two interfaces. Participants received instruction to hit the stop button if and only if they thought the arm was about to collide with a tower. Then, after a 3-2-1 countdown, we started the arm moving.

The monitor and MR interfaces then followed. We alternated their order across participants. Participants received instruction to label the robot's planned motion as quickly and accuracy as possible. Then, after a 3-2-1 countdown, we displayed the visualization. After completing the task with each interface, the participant completed three questionnaires. 

\subsection{Measurements}
We chose the choice of interface as the within-subjects independent variable. In all three interfaces, our objective dependent variables were the true positive rate of correctly classifying a path as colliding, and the true negative rate of correctly classifying a path as non-colliding. We also accounted for participant strategy in labeling each motion as colliding or non-colliding (e.g., showing a tendency to always label a motion plan as colliding). 

In the monitor and HoloLens interface conditions, we also measured the average speed of labeling each motion plan by recording the time elapsed from first seeing the visualization of the planned path to labeling the path. This allowed us to measure the accuracy and precision with which each interface allowed participants to label the robot's intended motion. 

Our subjective dependent variables were participant workload as measured by the NASA Task Load Index (NASA-TLX) questionnaire \cite{nasa1987}, system usability as measured by the System Usability Scale (SUS) questionnaire \cite{brooke96}, and our own questionnaire measuring perceived predictability and preference for each interface.
\begin{itemize}
\item \textbf{NASA-TLX:} This is a widely-used assessment questionnaire which asks participants to provide a rating of their perceived workload during a task across six sub-scales: mental demand, physical demand, temporal demand, effort, frustration, and performance. We measured the first five on scales from 0 (Low) to 100 (High), with performance measured from 0 (Perfect) to 100 (Failure). For this evaluation, the weighted measure of paired comparisons among the sub-scales was not included. The workload score is calculated as the average of the six sub-scales. 
\item \textbf{SUS:} This questionnaire assesses overall system usability by asking participants to rate ten statements on a 7-point Likert scale ranging from ``strongly disagree'' to ``strongly agree.'' The statements cover different aspects of the system, such as complexity, consistency, and cumbersomeness. SUS is measured on a scale from 0 to 100, where 0 is the worst score and 100 is the best.
\item \textbf{Ours:} This assessed how participants felt each interface helped them to accurately predict collisions. Participants were asked to rate three statements, one for each condition, on a 7-point Likert scale ranging from ``strongly disagree'' to ``strongly agree.'' For instance, ``When using the monitor and keyboard, I felt I could accurately predict collisions.'' In addition, we asked participants to select which interface they enjoyed the most, which interface made understanding the robot's motion the easiest, and which interface they preferred for completing the task.
\end{itemize}

\subsection{Hypotheses}
We expected that participants would show the best performance (i.e. highest true positives/negatives, least false positives/negatives, lowest levels of mental workload, highest usability, predictability, and system preference scores) in the Mixed Reality interface condition followed by the monitor interface. Additionally, we hypothesize that participants would have a faster labeling speed with the MR interface compared to the monitor interface. 

\begin{itemize}
\item \textbf{H1:} The HoloLens will be the easiest interface for completing the motion labeling task. This will be demonstrated by participants achieving the best performance out of the three conditions, across (a) highest true positives/negatives, (b) lowest false positives/negatives, (c) lowest levels of workload, (d) highest usability scores, and (e) highest predictability and preference scores. 
\item \textbf{H2:} The monitor interface will be easier for completing this task than using no visualization at all. This will be demonstrated by participants achieving better performance than with no visualization, across (a) higher true positives/negatives, (b) lower false positives/negatives, (c) lower levels of workload, (d) higher usability scores, and (e) higher predictability and preference scores.
\item \textbf{H3:} The HoloLens interface will be associated with quicker labeling times than the monitor interface, as demonstrated by the average time it took for participants to label each motion as colliding or not colliding. Labeling times in the monitor and MR conditions are a function of evaluating the visualization of the robot's planned motion, whereas in the no visualization condition, labeling times are generated by watching the robot enact the planned motion. As a result, only the monitor and MR conditions are directly comparable.
\end{itemize}

\subsection{Results}

\subsubsection*{Analysis Techniques}

We used repeated measures analysis of variance (ANOVA) and signal detection theory (SDT) to determine if differences between measures in the three conditions were significant at the 95\% confidence level. While ANOVA is likely to be familiar to the reader, SDT is less likely to be familiar, and so we will describe its use.

Signal detection theory (SDT) describes accuracy in human perception and decision making tasks by taking into account preferences for certain types of responses \cite{macmillan2002signal, tanner1954decision}. For instance, in our task, always responding that a motion plan will collide would yield high true positive scores (``hits''), and also high false positive scores (``false alarms''). In decision making tasks with innocuous false alarms, adopting such a strategy would not affect overall performance. However, for tasks with high false alarm cost, then a strategy that results in low false alarm rates while retaining high hit rates is better. For human-robot interaction tasks like ours, false alarms would slow the collaboration considerably and so we consider them high cost.

In SDT tasks, $d'$ (also called sensitivity) is a common measure which considers decision making strategy. It is the standardized difference between the hit rate and the false alarm rate. To handle perfect scores (i.e., correctly labeling all the colliding and non-colliding paths), zero false alarm scores, and zero hit scores, we adopted the technique outlined by Stanislaw and Todorov \cite{stanislaw1999calculation}.

\subsubsection*{Accuracy}

\begin{figure}[t]
\centering
\begin{subfigure}{.47\linewidth}
  \centering
  \includegraphics[width=\textwidth]{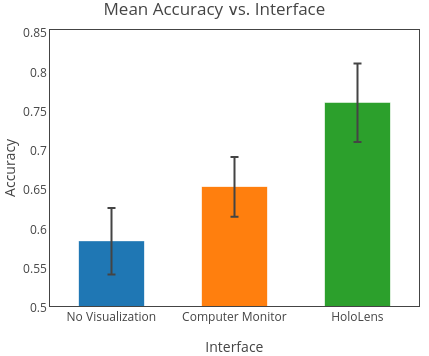}
  \caption{Mean accuracies across interfaces.}
  \label{fig:acc}
\end{subfigure}
\begin{subfigure}{.52\linewidth}
  \centering
  \includegraphics[width=\textwidth]{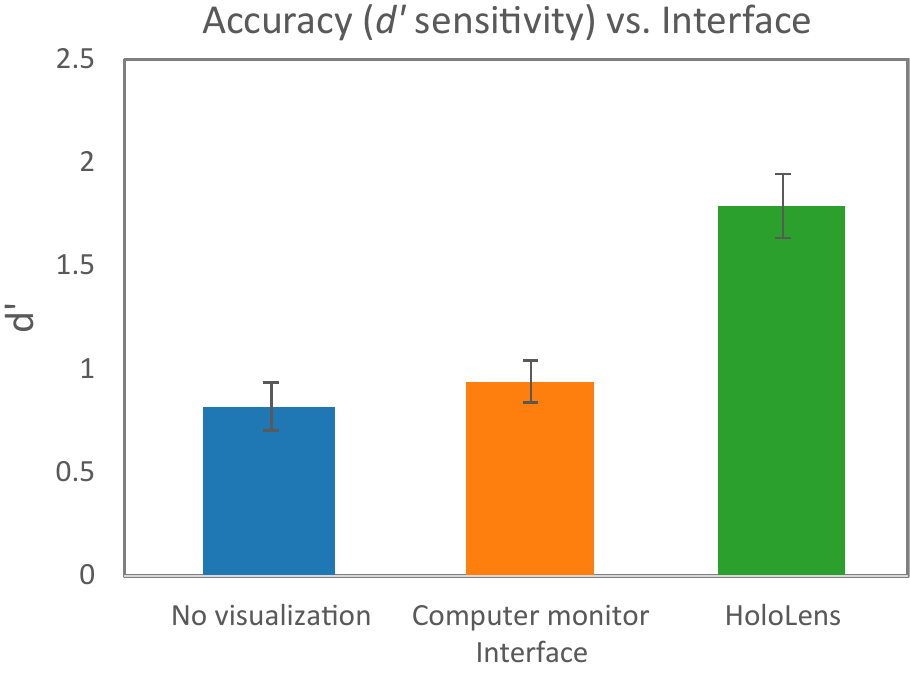}
  \caption{Mean adjusted accuracy ($d'$) across interfaces.}
  \label{fig:dprime}
\end{subfigure}
\\
\begin{subfigure}{.5\textwidth}
  \centering
  \includegraphics[width=\textwidth]{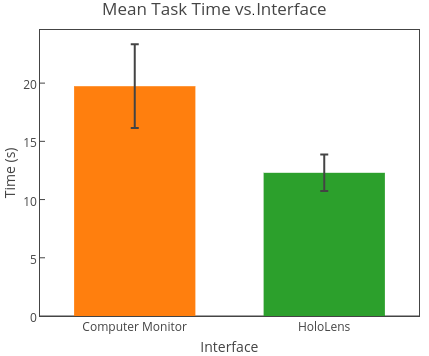}
  \caption{Mean task times for comparable interfaces (please see \textbf{H3} definition).}
  \label{fig:time}
\end{subfigure}%
\caption{The HoloLens interface significantly increases accuracy (\emph{top left}) and strategy-compensated accuracy ($d'$) (\emph{top right}) over the baseline systems. \emph{Bottom:} The HoloLens interface is also significantly faster for assessing motion plans.}
\label{fig:objective}
\end{figure}

We counted the number of participant true positives, false positives, true negatives, and false negatives in each condition. From this, we report the familiar accuracy measure as the proportion of true positives plus true negatives out of the total number of motion plans (Fig.~\ref{fig:objective}b). MR was the most accurate  $\mbox{(M= 0.76, SD= 0.19)}$, followed by the monitor $\mbox{(M= 0.66, SD= 0.14)}$, followed by the no visualization condition $\mbox{(M= 0.60, SD= 0.12)}$. These differences were statistically significant Wilks $\Lambda=0.619$,  $F(2,30)= 9.244$, $p=.001$, $\eta^2=0.381$, and accuracy in the MR condition was significantly better than in the monitor condition ($p=.001$) and the no visualization condition ($p<.001$). Performance in the monitor condition was not statistically significantly better than in the no visualization condition, $p=.065$.

We also report $d'$ scores for each participant in each of the three conditions (Fig.~\ref{fig:objective}c). We conducted a one-way repeated measures ANOVA to look for significant differences in $d'$ scores across the three conditions. There was a statistically significant difference in $d'$ performance scores between the conditions, Wilks $\Lambda=0.449$  $F(2,30)= 18.378$, $p<.001$, $\eta^2=0.551$. Further, the performance in the MR condition $\mbox{(M= 1.79, SD= 0.88)}$ was statistically significantly better than the monitor condition $\mbox{(M= 0.94, SD= 0.58)}$ and the no visualization condition ($M= 0.82, SD= 0.66$), all $p<.001$. The difference between performance in the monitor condition was not statistically significantly better than performance in the no visualization condition, $\mbox{p= .42}$. A look at the mean accuracy and mean $d'$ scores showed that performance in the MR, monitor, and no visualization conditions trended in the hypothesized direction although both performance indicators in the monitor condition were not statistically significantly better the no visualization condition. Thus, hypotheses 1 (a) and (b) were supported, but hypotheses and 2 (a) and 2 (b) were not supported.

\subsubsection*{Task Time}

Hypothesis 3 stated that motion labeling times would be faster in the HoloLens condition than in the monitor condition. We conducted a paired samples t-test, which showed statistically significant differences in mean motion labeling times between the two conditions ($t(31)=3.415, p<.001$). Mean labeling times trended in the hypothesized direction (Fig.~\ref{fig:time}). Labeling times in the HoloLens condition were significantly shorter ($M=11.95, SD=8.42$) than in the monitor condition ($M=19.39, SD=19.28$). Hypothesis 3 was supported.

\subsubsection*{Subjective Workload}

\begin{figure}[t]
\centering
\begin{subfigure}{.5\textwidth}
  \centering
  \includegraphics[width=\textwidth]{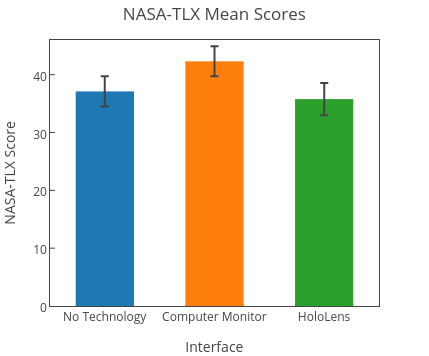}
  \caption{Mean NASA-TLX scores across all interfaces.}
  \label{fig:nasa}
\end{subfigure}%
\begin{subfigure}{.5\textwidth}
  \centering
  \includegraphics[width=\textwidth]{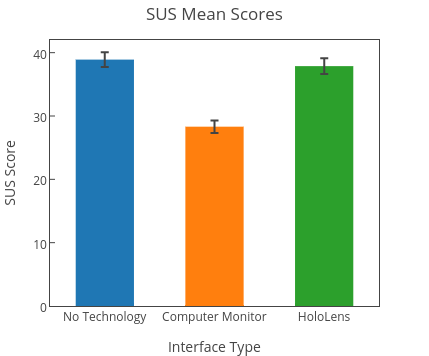}
  \caption{Mean SUS scores across all interfaces.}
  \label{fig:sus}
\end{subfigure}
\caption{Participants reported the lowest levels of subjective workload in the HoloLens condition and statistically significantly lower workload than in the monitor condition (NASA-LTX). Participants reported the highest assessments of system usability in the no visualization condition (SUS). However, the difference between the no visualization condition and the HoloLens condition was not statistically significant.}
\label{fig:subjective}
\end{figure}

Hypotheses 1 (c) and 2 (c) stated that the HoloLens would be associated with the lowest levels of subjective workload across the 3 interface conditions, and the monitor interface condition would be associated with lower levels of workload than the no visualization condition. To test this, we conducted one-way repeated measures ANOVA. There was a statistically significant difference in scores on the workload measure across the three interface conditions,  
Wilks $\Lambda=0.802$,  $F(2,30)= 3.693$, $p=0.037$, $\eta^2=0.198$ 

The HoloLens condition was associated with the lowest workload scores ($M=35.39, SD=2.78$), followed by the no visualization condition ($M=37.11, SD=2.61$), and then the monitor condition ($M=42.32, SD=14.71$). Post hoc comparisons showed that the mean scores in the HoloLens condition were statistically significantly lower than in the monitor condition $p=.040$. There was not a statistically significant difference in workload scores between the HoloLens condition and the no visualization condition. The difference between workload scores in the monitor condition and the no visualization condition were not significantly different. Hypotheses 1 (c), which stated that the HoloLens would have the lowest workload scores, was partially supported. Hypothesis 2 (c) was not supported as the workload scores in the monitor condition were higher than in the no visualization condition.

\subsubsection*{Subjective Usability}

Hypotheses 1 (d) and 2 (d) stated that usability scores in the HoloLens condition would be the highest of the three conditions and that usability scores in the monitor condition would be higher than in the no visualization condition. Using one-way repeated measures ANOVA, the results show that there was a statistically significant difference in mean usability scores across the three conditions, Wilks $\Lambda=0.151$, $F(2,30)= 84.342$, $p<0.001$, $\eta^2=0.849$. However, the no visualization condition was associated with the highest SUS scores ($M=38.91, SD=1.15$), followed by the HoloLens condition ($M=37.88, SD=1.26$), and the monitor condition ($M=28.31, SD=0.99$). Mean SUS scores in the HoloLens condition were significantly higher than the monitor condition, $p<0.001$, and mean SUS scores in the no visualization condition were significantly higher than in monitor condition, $p<0.001$. The difference between the HoloLens condition and the no visualization condition was not significant. Hypotheses 1 (d) and 2 (d) were not supported.

\subsubsection*{Subjective Collision Predictability}

Hypotheses 1 (e) and 2 (e) stated that the HoloLens condition would be associated with the highest collision predictability scores of the three conditions, and that predictability scores in the monitor condition would be higher than the no visualization condition. We used one-way repeated measures ANOVA to test for statistically significant differences in participants' assessments of whether or not they felt the interfaces could help them predict collisions. There were statistically significant differences between the interfaces on whether or not participants felt the interface could help them accurately predict collisions, Wilks $\Lambda=0.246$  $F(2,30)= 45.891$, $p<0.001$, $\eta^2=0.754$ participants showed the highest agreement that the HoloLens helped them to accurately predict collisions ($M=5.28, SD=0.20$), followed by the no visualization condition ($M=4.06, SD=0.35$) and then the monitor condition ($M=3.38, SD=0.23$). Further, the difference between mean scores in the HoloLens condition were significantly higher than in the monitor condition,and the no visualization condition (both $p's<.05$), supporting Hypothesis 1 (e). Means scores in the monitor condition were lower than the no visualization condition but not significantly so ($p=.44$). Hypothesis 2 (e) was not supported. 

\subsubsection*{Subjective Enjoyment}

We compared the frequency with which participants selected each interface as the one they enjoyed the most, the one they preferred for completing the task, and the one they felt made understanding the robot's motion the easiest. All participants selected the HoloLens as the interface they enjoyed the most ($N=32$). For the interface participants felt made understanding the robot's motion the easiest, almost all of the participants selected HoloLens ($N=29, 90.6\%$), while only three participants ($9.4\%$) selected the monitor. Finally, when asked about preference for completing that task, almost all participants selected the HoloLens ($N=30, 93.8\%$). Only two participants ($6.3\%$) selected the monitor interface as their preferred interface for completing the task. No participants selected the no visualization condition.

\section{Discussion}

Overall, our results demonstrate the potential benefit of MR to communicate robot motion intent to humans. Participants in the MR condition significantly outperformed the monitor condition, showing a 16\% increase in collision prediction accuracy and a 62\% decrease in time taken. Mixed reality also allowed participants to outperform the control condition of no visualization. Almost universally, participants selected the HoloLens as the most enjoyable interface, the easiest for completing the task, and the one they preferred for assessing the robot motion plans. Taken together, these findings strongly support our hypotheses that MR would be associated with the best objective performance measures. 

An examination of participant free responses regarding why they preferred MR over monitor offers some insight into these findings. Many participants reported that using the monitor and mouse to virtually move around the robot was cumbersome, unintuitive, difficult to manipulate, distracting, and confusing. Participants reported that MR was not perfect, e.g., the motion plan overlay was not always perfectly aligned on top of the robot, the setup took a long time, and that physically moving around the robot was difficult at times. Even so, 34\% of participants reported that they liked that they could freely move around the robot to see the planned motion, and that this made determining whether or not collisions would occur faster, easier, and more intuitive than when using the monitor and mouse.

The subjective questionnaire responses offered mixed but promising support for the MR condition. Although participants working with the MR condition reported lower workload than in the no visualization condition, it was not significantly lower, which offered only partial support for hypothesis H1 (c). The mean workload scores did trend in the hypothesized direction (i.e., the MR condition had the lowest workload scores overall) and the results imply that participants found working with the MR interface no more taxing than using no interface at all for this task. Similarly, although participants rated the no visualization condition as slightly more usable than the MR condition (counter to hypothesis H1 (d)), the no visualization condition was not rated significantly more usable. It is promising to see that wearing the MR was perceived to have as low a mental demand and to be as easy as simply watching the robot move through the environment with no HMD. 

Perhaps surprisingly, the monitor condition did not significantly outperform the no visualization condition for both objective and subjective measures. Participant accuracy (and accuracy accounting for decision making strategy) was not significantly better, and when working with the computer monitor, participants reported higher workload and lower assessments of usability than when working with the no visualization condition. Put another way, looking at a robot with an emergency stop button in your hand is about as simple an interface as you could build. Finally, participants also reported the least agreement that the monitor interface could help them to accurately predict robot collisions. Thus, no part of hypothesis 2 was supported.

\subsection{Limitations}

At a high level, our system only considers robot-to-human motion intention communication, and there is much to explore still in enabling human-to-robot communication. Further, it only considers visual communication of robot motion states, and this could be expanded to non-visual communication and communication of non-motion states. Mixed reality can also be used to communicate shared goals, to indicate which objects are to be manipulated, and to enable cooperative behaviors. 

At the system level, the HoloLens field of view for graphical display is small. Further, although the HoloLens has the ability to create a spatial map of the environment, it does not occlude virtual objects when they are placed behind real objects.

\section{Conclusion}

If robots and humans are to form fluid cooperative work partnerships, then they need to be able to communicate their motion intent to each other effectively. We investigated the speculation that mixed reality would be a natural interface for robot motion intent communication, and found that both participant performance and participant perceptions overwhelmingly supported an MR visualization over the more traditional monitor interface for visualization and over no visualization at all. Our results provide evidence that mixed reality is one way to bridge the communication gap and allow robots to communicate their motion intent to humans.

\begin{acknowledgement}
We thank David Laidlaw for fruitful discussion on the virtual reality literature. This material is based upon work supported by the Defense Advanced Research Projects Agency (DARPA) under grant number D15AP00102 and by the Air Force Research Laboratory (AFRL) under grant number FA9550-17-1-0124. Any opinions, findings, and conclusions or recommendations expressed in this material are those of the authors and do not necessarily reflect the views of DARPA or AFRL.
\end{acknowledgement}

\vfill
\newpage

\bibliographystyle{abbrv}
\bibliography{bibliography}

\end{document}